\pdfoutput=1

\documentclass[11pt]{article}

\usepackage[preprint]{acl}

\usepackage{times}
\usepackage{latexsym}

\usepackage[T1]{fontenc}

\usepackage[utf8]{inputenc}

\usepackage{microtype}

\usepackage{inconsolata}

\usepackage{graphicx}

\usepackage{tabularx}
\usepackage{multirow}
\usepackage{booktabs}
\usepackage{amsfonts}
\usepackage{pifont}
\usepackage{wrapfig}

\usepackage{float}
\restylefloat{table}

\usepackage[skip=4pt]{caption}  
\usepackage{enumitem}
\newlist{todolist}{itemize}{2}
\setlist[todolist]{label=$\square$}
\usepackage{pifont}

\usepackage{cleveref}
\title{SQuARE: Sequential Question Answering Reasoning Engine for Enhanced Chain-of-Thought in Large Language Models}
\newcommand{\rephrase}{SQuARE}

\author{Daniel Fleischer\thanks{Corresponding author: \href{mailto:daniel.fleischer@intel.com}{daniel.fleischer@intel.com}} 
\qquad Moshe Berchansky \qquad Gad Markovits \qquad Moshe Wasserblat
\vspace{0.3em} \\
Intel Labs
}

\begin{document}
\maketitle

\begin{abstract}

In the rapidly evolving field of Natural Language Processing, Large Language Models (LLMs) are tasked with increasingly complex reasoning challenges. Traditional methods like chain-of-thought prompting have shown promise but often fall short in fully leveraging a model's reasoning capabilities. This paper introduces \rephrase{} (Sequential Question Answering Reasoning Engine), a novel prompting technique designed to improve reasoning through a self-interrogation paradigm. Building upon CoT frameworks, \rephrase{} prompts models to generate and resolve multiple auxiliary questions before tackling the main query, promoting a more thorough exploration of various aspects of a topic. Our expansive evaluations, conducted with Llama 3 and GPT-4o models across multiple question-answering datasets, demonstrate that \rephrase{} significantly surpasses traditional CoT prompts and existing rephrase-and-respond methods. By systematically decomposing queries, \rephrase{} advances LLM capabilities in reasoning tasks. The code is publicly available at \url{https://github.com/IntelLabs/RAG-FiT/tree/square}.

\end{abstract}

\begin{table*}[ht]
    \caption{Main prompt for the \rephrase{} technique. 
    }
    \centering
    \small
    \begin{tabular}{>{\raggedright\arraybackslash\tt}p{0.98\textwidth}<{}}
      \toprule
      You are a helpful question answerer who can provide an answer given a question and relevant context.\\
      Generate \{N\} questions based on the given question and context, and shortly answer them.\\
      Finally, provide an answer to the original question using what you learned from answering the questions you created.
      The answer should be a short span, just a few words. \\
      \bottomrule
    \end{tabular}
    \label{tab:square_prompt}
\end{table*}

\section{Introduction}
\label{sec:introduction}

Large Language Models (LLMs) have rapidly transformed Natural Language Processing (NLP), excelling in tasks like text generation, machine translation, and dialogue systems \cite{brownLanguageModelsAre2020,kojimaLargeLanguageModels2022}. These models owe their flexibility to the Transformer architecture \cite{vaswaniAttentionAllYou2017}, and benefit from large-scale pretraining followed by fine-tuning or instruction tuning to align with human objectives \cite{ouyangTrainingLanguageModels2022,weiFinetunedLanguageModels2022}. A key technique for enhancing these models is chain-of-thought (CoT) prompting, which has gained notable attention for its ability to improve reasoning by encouraging models to work through problems step by step \citep{weiChainofThoughtPromptingElicits2023}. This approach has shown efficacy in complex tasks like multi-step arithmetic and commonsense question answering, by making intermediate processes transparent and facilitating more accurate outcomes \cite{snellScalingLLMTestTime2024}. While some CoT variants explore iterative reasoning, there is still limited exploration of self-interrogation paradigms that prompt models to pose and resolve their own intermediate queries.

\begin{figure}[t]
    \centering
    \includegraphics[scale=0.9]{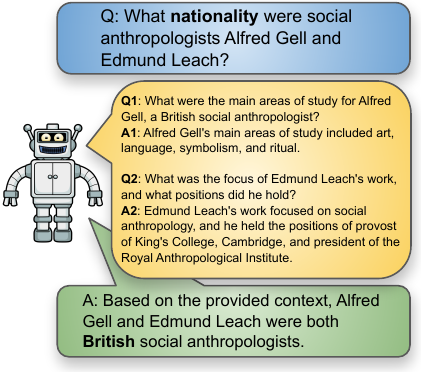}
    \caption{The \rephrase{} methods prompts the model to generate $N$ question-answer pairs about the topic and then respond to the original query, having established additional context.}
    \label{fig:illustration}
\end{figure}

In this paper, we introduce \rephrase{} (Sequential Question Answering Reasoning Engine), a prompting technique that instructs an LLM to generate and answer multiple sub-questions before addressing the main query. By decomposing queries into iterative steps, \rephrase{} draws on chain-of-thought frameworks and prior prompting methodologies~\citep{dengRephraseRespondLet2024} to produce more comprehensive solutions. In extensive evaluations on multiple question-answering datasets using Llama 3~\cite{grattafioriLlama3Herd2024} (3B and 8B) and \mbox{GPT-4o} \citep{openaiGPT4oSystemCard2024}, \mbox{\rephrase{} outperforms} chain-of-thought prompts and existing rephrase-and-respond strategies. This work highlights how systematically breaking down queries advances LLM reasoning capabilities.

\section{\rephrase}
\label{sec:phrasing}

In this section, we introduce the \rephrase{} technique in more detail. Building upon the foundation laid by \citet{dengRephraseRespondLet2024}, our method alters the system instructions to prompt the model to generate a set of $N$ question-and-answer pairs. Figure~\ref{fig:illustration} illustrates a simple example in which the model receives a query, generates two sub-questions and their corresponding answers, and then arrives at a correct final solution. The system prompt used by our method is presented in Table~\ref{tab:square_prompt}.

The rationale behind \rephrase{} is to guide the model into an iterative cycle of inquiry and response, encouraging it to explore various facets of a topic before forming a conclusion. In contrast to standard chain-of-thought prompts, which often present a single stream of reasoning, \rephrase{} nudges the model toward self-interrogation pathways. This design also makes \rephrase{} relatively straightforward to integrate with other prompting techniques. In practice, $N$ can be tuned to balance the thoroughness of exploration with computational cost and response length; our experiments in Section~\ref{sec:experiments} show that even a small set of sub-questions can significantly improve the final answers' correctness.

\begin{table*}[ht]
    \centering
    \caption{The main results of our experimentation. Each row group corresponds to the results for the given dataset, with each row showcasing the metric results for each model. The columns include all the main approaches, with \textbf{bold} highlighting the best result across all approaches.}
    \small
    \begin{tabular}{llccccc}
      \toprule
      Dataset & Model & Baseline & RAG & CoT & RaR & \rephrase \\
      \midrule
      \multirow[l]{3}{*}{TriviaQA}
          & Llama-3.2 3B  & 59.5 & 82.0 & 87.5  & 86.0 &  \textbf{88.5}    \\
          & Llama-3.1 8B  & 76.5 & 89.5 & 90.5  & 89.5 &  \textbf{92.5}    \\
          & GPT-4o    & 88.7 & 92.7 & 92.7  & 94.7 &  \textbf{96.7}    \\
      \midrule
      \multirow[l]{3}{*}{HotpotQA}
          & Llama-3.2 3B  &  17.5  & 26.0  & 26.5   & 25.0  &  \textbf{31.5}   \\
          & Llama-3.1 8B  &  23.0  & 26.5  & 31.0   & 28.5  &  \textbf{33.5}   \\
          & GPT-4o    &  44.0  & 45.3  & 46.7   & \textbf{47.3}  &  46.7   \\
      \midrule
      \multirow[l]{3}{*}{ASQA}
          & Llama-3.2 3B  &  14.2 & 21.5  & 21.9  & 23.5  &  \textbf{26.6}   \\ 
          & Llama-3.1 8B  &  14.6 & 23.1  & 24.8  & 25.5  &  \textbf{28.8}   \\ 
          & GPT-4o    &  26.8 & 30.4  & \textbf{31.9}  & 30.1 & 31.7 \\ 
      \bottomrule
    \end{tabular}
    \label{tab:main}
\end{table*}

\section{Experiments}
\label{sec:experiments}

In this section, we detail the experimental setup and the evaluations conducted to assess the effectiveness of the \rephrase{} technique across various datasets and models. Our approach is compared to several existing methods to ascertain its relative performance.

\subsection{Datasets}

We evaluate our models on \mbox{\textbf{TriviaQA}} \cite{joshi-etal-2017-triviaqa}, \textbf{HotpotQA} \cite{yang-etal-2018-hotpotqa}, and \textbf{ASQA} \cite{stelmakh-etal-2022-asqa} which are knowledge intensive question-answering datasets which benefit from external context. Context retrieval was done over a Wikipedia corpus \cite{zhangRetrieveAnythingAugment2023}. We randomly sampled 200 examples from each dataset.  Results are reported using the following metrics: for TriviaQA and HotpotQA sub-string exact match (subEM) is reported \cite{asaiSelfRAGLearningRetrieve2023,yenHELMETHowEvaluate2024}. For ASQA, recall-EM is reported \cite{gaoEnablingLargeLanguage2023a}. For more details, see \Cref{sec:datasets}.

\subsection{Models}
\label{sec:models}

Our experiments utilize two open-source Llama models \cite{grattafioriLlama3Herd2024}: Llama-3.2 3B and Llama-3.1 8B. Both models are instruction-tuned to optimize their performance on complex tasks. In addition, we employed the OpenAI GPT-4o system\footnote{Version \textit{2024-05-13}.} \cite{openaiGPT4oSystemCard2024} to provide a benchmark for comparison. We use greedy decoding with local models.

\subsection{Configurations}
\label{sec:configs}

Our experimental setup is composed of the following configuration settings:

\begin{itemize}[noitemsep,topsep=0.5em,parsep=0.4em,leftmargin=0.8em]
\item \textbf{Baseline}: Standard application without any augmentative techniques.
\item \textbf{CoT}: Methodology as outlined by \citet{weiChainofThoughtPromptingElicits2023} that leverages intermediate reasoning steps leading to a final answer; instruction described in \Cref{tab:cot_prompt}.
\item \textbf{RaR:} A rephrasing strategy that prompts for a rephrasing of the original request before answering it, as proposed by \citet{dengRephraseRespondLet2024}; instruction described in \Cref{tab:rar_prompt}.
\item \textbf{\rephrase}: This configuration employs our prompt and is run with a default $N\!=\!3$ question-answer pairs.
\end{itemize}

We augment the requests with a pair of query-answer examples (few-shot) to facilitate understanding and improve prediction formatting and accuracy. All prompts and few-shot examples are presented in \Cref{sec:prompts} for reproducibility.

\pagebreak

Notably, in configurations containing reasoning instructions, we employ a regular expression\footnote{Regex pattern: \texttt{.*answer(.*)}. It has a 99.2\% capture rate.} to extract the final answer. This extraction is crucial as it assists in mitigating incorrect answers when correct phrases appear throughout reasoning chains \textbf{but not in the final answer}. For an example of this phenomena, see \Cref{tab:bad-example}.

\section{Results}
\label{sec:results}

Table~\ref{tab:main} presents the main results of our method compared against several baselines on three benchmark QA datasets: TriviaQA, HotpotQA, and ASQA. Across the smaller Llama 3.2 3B and Llama 3.1 8B models, our approach consistently outperforms or matches the strongest baselines in each dataset. For example, with Llama 3.2 3B on \mbox{TriviaQA}, \rephrase{} improves performance by 6.5\% and 2.5\% over RAG and RaR, respectively, achieving an overall score of 88.5\%. On HotpotQA, Llama 3.2 3B also sees a notable boost, from 26.5\% (CoT) to 31.5\% with our method. These gains become even more pronounced with Llama 3.1 8B, where improvements of up to 3\% (TriviaQA) and 7\% (HotpotQA) are observed compared to alternative methods. We also observe notable gains on ASQA. For Llama-3.2 3B, \rephrase{} lifts performance from 21.5\% (RAG) and 23.5\% (RaR) to 26.6\%, nearly doubling the baseline of 14.2\%. 

When using GPT-4o, \rephrase{} remains highly competitive. On TriviaQA, our method reaches 96.7\%, outperforming other settings by at least 2.0\%. On HotpotQA, RaR and \rephrase{} are close, with RaR exhibiting a slight edge (47.3\% versus 46.7\%). For ASQA, CoT and \rephrase{} yield nearly identical performance (31.9\% versus 31.7\%), indicating that GPT-4o is already adept at leveraging additional reasoning steps or retrieved facts in these tasks. Nevertheless, \rephrase{} demonstrates robust performance across all three datasets and is especially beneficial for smaller-scale models, where sequential questioning can substantially bolster the final answer quality.

\begin{table}[t]
    \centering
    \small
    \begin{tabular}{llccc}
      \toprule
      Dataset & $N$ &  \rephrase & +Summarize  & +Vote \\
      \midrule
      \multirow[l]{3}{*}{TriviaQA}
              & 3  & 92.5 & 87.5 & 81.0  \\
              & 5  & \textbf{94.0} & 85.5 & 78.0  \\
              & 10 & \textbf{94.0} & 88.0 & 89.0  \\
      \midrule
      \multirow[l]{3}{*}{HotpotQA}
              & 3  & \textbf{33.5} & 30.0 & 23.5    \\
              &  5 & 31.5 & 31.5 & 22.5    \\
              &  10 & \textbf{33.5} & 29.0 & 23.5    \\
      \midrule
      \multirow[l]{3}{*}{ASQA}
              & 3 & \textbf{28.8} & 20.9 & 23.9   \\ 
              & 5 & 27.9 & 22.1 & 23.5  \\ 
              & 10 & 27.8 & 23.1 & 22.7  \\ 
      \bottomrule
    \end{tabular}
    \caption{Comparison of two aggregation methods in addition to \rephrase{}, and the effect of varying the number of sub-questions ($N$). Results showcase the Llama-3.1 8B model with few-shot examples adapted for each approach, as detailed in \Cref{sec:prompts}.}
    \label{tab:variations}
\end{table}

\subsection{Ablation Study}
\label{sec:ablation-study}

To highlight the contribution of each component in \rephrase, we performed an ablation study analyzing (1) the number of generated questions ($N$), (2) the role of few-shot examples, and (3) an optional aggregation step.

\paragraph{Number of Generated Questions:}
We conducted an evaluation using $N\in\{3,5,10\}$. As shown in Table~\ref{tab:variations}, for TriviaQA, increasing $N$ from 3 to 5 or 10 boosts performance from 92.5\% to 94.0\%. On HotpotQA, $N\!=\!5$ (31.5\%) dips slightly below $N\!=\!3$, but returns to 33.5\% at $N\!=\!10$. In ASQA, performance drops from 28.8\% at $N\!=\!3$ to 27.8\% at $N\!=\!10$, suggesting that while additional questions can add useful context, they can also introduce redundancy or noise. For more comparisons, see \Cref{tab:more_results}.

\begin{figure}[t]
    \centering
    \includegraphics[scale=0.168]{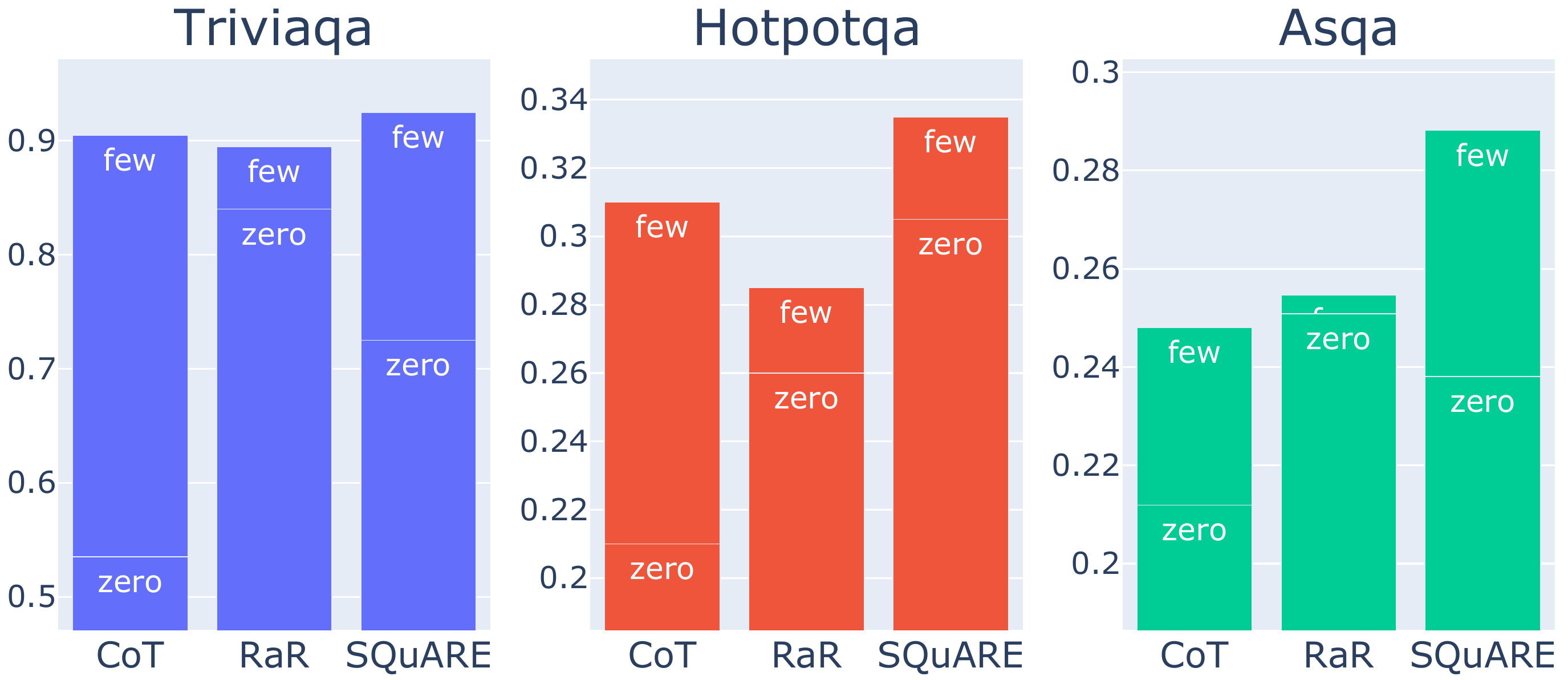}
    \caption{Ablation study illustrating how few-shot examples influence performance metrics for the CoT, RaR, and \rephrase{} approaches, using the Llama-3.1 8B model.}
    \label{fig:fewshot}
\end{figure}


\paragraph{Impact of Few-Shot Examples:}
We inspected how incorporating few-shot examples substantially boosts accuracy, as seen in Figure~\ref{fig:fewshot}. We observe that both CoT and \rephrase{} benefit strongly from these examples, indicating that better exposure to task-relevant scenarios helps the model generate answers with correct and properly formed final answers. Interestingly, zero-shot experiments exhibit lower regex capture rate (85.0\%, see \Cref{sec:configs}) which could play a role in the diminished performance. For full results, see \Cref{tab:more_results}.

\paragraph{Aggregation Methods:}
Finally, we explored two aggregation strategies, before producing the final answer: \textit{Summarize} and \textit{Vote}. The \textit{Summarize} method involves the model summarizing the information learned from the generated questions and answers, whereas the \textit{Vote} method relies on majority voting to determine the final answer. According to \Cref{tab:variations}, \textit{Summarize} generally outperforms \textit{Vote} on TriviaQA and HotpotQA. However, using no aggregation step outperforms both in nearly all instances, suggesting that further post-processing can sometimes hurt the quality of the answer.

\section{Related Works}
\label{sec:related-works}

Chain-of-Thought (CoT) prompting, introduced by \citet{wu-etal-2023-chain}, and explored further by \citet{weiChainofThoughtPromptingElicits2023}, has been instrumental in enhancing language models, by encouraging them to articulate their reasoning processes explicitly. This approach has been shown to substantially improve model performance across a wide range of tasks, including question answering. 

\citet{dengRephraseRespondLet2024} propose a novel rephrasing prompt, which involves requesting the model to rephrase the initial question before providing an answer. This method has demonstrated performance improvements on various datasets, highlighting its efficacy in refining model responses. Our work expands upon this approach, by utilizing multiple query-answer pairs, that enable the model to better examine the topic at hand, and provide a better answer.

\citet{wangSelfConsistencyImprovesChain2023} and \citet{chenUniversalSelfConsistencyLarge2023} leverage self-consistency techniques by generating multiple response samples (by using sample decoding) and incorporating an aggregation step to increase accuracy, thereby enhancing the reliability of model conclusions. While our approach does generate multiple variations of the possible answer, they are dedicated for answering specific automatically generated inquiries regarding the topic at hand.

\citet{snellScalingLLMTestTime2024} demonstrate that extra test-time compute boosts LLM performance on difficult prompts, with smaller models sometimes surpassing larger ones. They propose a \textit{compute-optimal} method that adaptively explores multiple next steps, maximizing inference efficiency. Building on this idea, our approach focuses on question answering, where diverse perspectives substantially improve response quality. As previously mentioned, while our approach benefits from generating multiple responses for a given query, we focus on specific query-answer pair generation.   

\section{Conclusions and Summary}
\label{sec:conclusions-summary}

This study introduced a multi-question chain-of-thought prompt strategy that significantly enhances the reasoning capabilities of large language models. By generating and answering a series of sub-questions before addressing the primary query, our method improves response accuracy over traditional baselines and established techniques such as canonical chain-of-thought and RaR \cite{dengRephraseRespondLet2024}. Experiments with Llama 3 models and GPT-4o on several Q\&A datasets show that our approach outperforms existing methods, highlighting its effectiveness.

These results show how carefully designed prompts can improve multi-step reasoning in large language models. They also point to the value of exploring adaptive prompt techniques across different NLP tasks. As these models evolve, multi-question prompting may further sharpen automated reasoning and foster more dependable AI interactions.

\section*{Limitations and Future Plans}

While our multi-question chain of thought prompt strategy has demonstrated notable improvements in reasoning capabilities and response accuracy of large language models, several limitations should be acknowledged. Firstly, the method requires fine-tuning of the number of intermediate questions (3, 5, 10 or other), and this may not be optimal or applicable across varying query complexities or domains. Choosing the appropriate number of questions is important, as an incorrect configuration might lead to redundancy or insufficient exploration of the query context.

Secondly, our approach was evaluated only on specific Q\&A datasets, which may not encompass the full spectrum of topics and question types. Therefore, the generalizability of this technique to other domains, such as dialogue systems or more complex multi-turn interactions, remains to be tested. Additionally, while our experiments utilized the Llama 3 models and GPT-4o, the effectiveness of this strategy across other architectures or smaller-scale models could differ.

Another limitation is the potential increase in computational resources required to generate and answer multiple intermediate questions, which could impact the efficiency and scalability of deploying these models in real-time applications. 


Future research should focus on addressing these limitations by exploring adaptive mechanisms for intermediate question generation, extending validation across more diverse datasets and models, and optimizing computational requirements to ensure broader applicability and effectiveness.

\section*{Ethics Statement}

Throughout our research, we carefully considered the ethical aspects of developing advanced language models. Our technique aims to enhance reasoning and accuracy, but we recognize the need to address potential ethical issues.

One concern is that improved reasoning could result in producing more persuasive but misleading or harmful content. To counteract this, it is essential to implement safeguards ensuring responses are accurate, unbiased, and factual. Future efforts should continue to monitor outputs for bias and misinformation, incorporating methods to mitigate these risks.

Additionally, the increased computational demand for generating intermediate questions raises environmental concerns about energy consumption. We advocate for continued research into optimizing the efficiency of these processes to minimize ecological impact.

We prioritized privacy and security by using only publicly available data in our experiments, free of private information. Adhering to transparency and reproducibility principles, we documented our methodologies, to facilitate replication of our findings by others.

\bibliography{anthology,library}

\begin{thebibliography}{20}
\providecommand{\natexlab}[1]{#1}

\bibitem[{Asai et~al.(2023)Asai, Wu, Wang, Sil, and
  Hajishirzi}]{asaiSelfRAGLearningRetrieve2023}
Akari Asai, Zeqiu Wu, Yizhong Wang, Avirup Sil, and Hannaneh Hajishirzi. 2023.
\newblock \href {https://doi.org/10.48550/arXiv.2310.11511} {Self-{{RAG}}:
  {{Learning}} to {{Retrieve}}, {{Generate}}, and {{Critique}} through
  {{Self-Reflection}}}.
\newblock \emph{Preprint}, arXiv:2310.11511.

\bibitem[{Brown et~al.(2020)Brown, Mann, Ryder, Subbiah, Kaplan, Dhariwal,
  Neelakantan, Shyam, Sastry, Askell, Agarwal, {Herbert-Voss}, Krueger,
  Henighan, Child, Ramesh, Ziegler, Wu, Winter, Hesse, Chen, Sigler, Litwin,
  Gray, Chess, Clark, Berner, McCandlish, Radford, Sutskever, and
  Amodei}]{brownLanguageModelsAre2020}
Tom~B. Brown, Benjamin Mann, Nick Ryder, Melanie Subbiah, Jared Kaplan,
  Prafulla Dhariwal, Arvind Neelakantan, Pranav Shyam, Girish Sastry, Amanda
  Askell, Sandhini Agarwal, Ariel {Herbert-Voss}, Gretchen Krueger, Tom
  Henighan, Rewon Child, Aditya Ramesh, Daniel~M. Ziegler, Jeffrey Wu, Clemens
  Winter, Christopher Hesse, Mark Chen, Eric Sigler, Mateusz Litwin, Scott
  Gray, Benjamin Chess, Jack Clark, Christopher Berner, Sam McCandlish, Alec
  Radford, Ilya Sutskever, and Dario Amodei. 2020.
\newblock \href {https://doi.org/10.48550/arXiv.2005.14165} {Language
  {{Models}} are {{Few-Shot Learners}}}.
\newblock \emph{Preprint}, arXiv:2005.14165.

\bibitem[{Chen et~al.(2023)Chen, Aksitov, Alon, Ren, Xiao, Yin, Prakash,
  Sutton, Wang, and Zhou}]{chenUniversalSelfConsistencyLarge2023}
Xinyun Chen, Renat Aksitov, Uri Alon, Jie Ren, Kefan Xiao, Pengcheng Yin,
  Sushant Prakash, Charles Sutton, Xuezhi Wang, and Denny Zhou. 2023.
\newblock \href {https://doi.org/10.48550/ARXIV.2311.17311} {Universal
  {{Self-Consistency}} for {{Large Language Model Generation}}}.

\bibitem[{Deng et~al.(2024)Deng, Zhang, Chen, and
  Gu}]{dengRephraseRespondLet2024}
Yihe Deng, Weitong Zhang, Zixiang Chen, and Quanquan Gu. 2024.
\newblock \href {https://doi.org/10.48550/arXiv.2311.04205} {Rephrase and
  {{Respond}}: {{Let Large Language Models Ask Better Questions}} for
  {{Themselves}}}.
\newblock \emph{Preprint}, arXiv:2311.04205.

\bibitem[{Gao et~al.(2023)Gao, Yen, Yu, and
  Chen}]{gaoEnablingLargeLanguage2023a}
Tianyu Gao, Howard Yen, Jiatong Yu, and Danqi Chen. 2023.
\newblock \href {https://doi.org/10.18653/v1/2023.emnlp-main.398} {Enabling
  {{Large Language Models}} to {{Generate Text}} with {{Citations}}}.
\newblock In \emph{Proceedings of the 2023 {{Conference}} on {{Empirical
  Methods}} in {{Natural Language Processing}}}, pages 6465--6488, Singapore.
  Association for Computational Linguistics.

\bibitem[{Grattafiori et~al.(2024)Grattafiori, Dubey, Jauhri, Pandey, Kadian,
  {Al-Dahle}, Letman, Mathur, Schelten, Vaughan, Yang, Fan, Goyal, Hartshorn,
  Yang, Mitra, Sravankumar, Korenev, Hinsvark, Rao, Zhang, Rodriguez,
  Gregerson, Spataru, Roziere, Biron, Tang, Chern, Caucheteux, Nayak, Bi,
  Marra, McConnell, Keller, Touret, Wu, Wong, Ferrer, Nikolaidis, Allonsius,
  Song, Pintz, Livshits, Wyatt, Esiobu, Choudhary, Mahajan, {Garcia-Olano},
  Perino, Hupkes, Lakomkin, AlBadawy, Lobanova, Dinan, Smith, Radenovic,
  Guzm{\'a}n, Zhang, Synnaeve, Lee, Anderson, Thattai, Nail, Mialon, Pang,
  Cucurell, Nguyen, Korevaar, Xu, Touvron, Zarov, Ibarra, Kloumann, Misra,
  Evtimov, Zhang, Copet, Lee, Geffert, Vranes, Park, Mahadeokar, Shah, van~der
  Linde, Billock, Hong, Lee, Fu, Chi, Huang, Liu, Wang, Yu, Bitton, Spisak,
  Park, Rocca, Johnstun, Saxe, Jia, Alwala, Prasad, Upasani, Plawiak, Li,
  Heafield, Stone, {El-Arini}, Iyer, Malik, Chiu, Bhalla, Lakhotia,
  {Rantala-Yeary}, van~der Maaten, Chen, Tan, Jenkins, Martin, Madaan, Malo,
  Blecher, Landzaat, de~Oliveira, Muzzi, Pasupuleti, Singh, Paluri, Kardas,
  Tsimpoukelli, Oldham, Rita, Pavlova, Kambadur, Lewis, Si, Singh, Hassan,
  Goyal, Torabi, Bashlykov, Bogoychev, Chatterji, Zhang, Duchenne, {\c C}elebi,
  Alrassy, Zhang, Li, Vasic, Weng, Bhargava, Dubal, Krishnan, Koura, Xu, He,
  Dong, Srinivasan, Ganapathy, Calderer, Cabral, Stojnic, Raileanu, Maheswari,
  Girdhar, Patel, Sauvestre, Polidoro, Sumbaly, Taylor, Silva, Hou, Wang,
  Hosseini, Chennabasappa, Singh, Bell, Kim, Edunov, Nie, Narang, Raparthy,
  Shen, Wan, Bhosale, Zhang, Vandenhende, Batra, Whitman, Sootla, Collot,
  Gururangan, Borodinsky, Herman, Fowler, Sheasha, Georgiou, Scialom,
  Speckbacher, Mihaylov, Xiao, Karn, Goswami, Gupta, Ramanathan, Kerkez,
  Gonguet, Do, Vogeti, Albiero, Petrovic, Chu, Xiong, Fu, Meers, Martinet,
  Wang, Wang, Tan, Xia, Xie, Jia, Wang, Goldschlag, Gaur, Babaei, Wen, Song,
  Zhang, Li, Mao, Coudert, Yan, Chen, Papakipos, Singh, Srivastava, Jain,
  Kelsey, Shajnfeld, Gangidi, Victoria, Goldstand, Menon, Sharma, Boesenberg,
  Baevski, Feinstein, Kallet, Sangani, Teo, Yunus, Lupu, Alvarado, Caples, Gu,
  Ho, Poulton, Ryan, Ramchandani, Dong, Franco, Goyal, Saraf, Chowdhury,
  Gabriel, Bharambe, Eisenman, Yazdan, James, Maurer, Leonhardi, Huang, Loyd,
  Paola, Paranjape, Liu, Wu, Ni, Hancock, Wasti, Spence, Stojkovic, Gamido,
  Montalvo, Parker, Burton, Mejia, Liu, Wang, Kim, Zhou, Hu, Chu, Cai, Tindal,
  Feichtenhofer, Gao, Civin, Beaty, Kreymer, Li, Adkins, Xu, Testuggine, David,
  Parikh, Liskovich, Foss, Wang, Le, Holland, Dowling, Jamil, Montgomery,
  Presani, Hahn, Wood, Le, Brinkman, Arcaute, Dunbar, Smothers, Sun, Kreuk,
  Tian, Kokkinos, Ozgenel, Caggioni, Kanayet, Seide, Florez, Schwarz, Badeer,
  Swee, Halpern, Herman, Sizov, Guangyi, Zhang, Lakshminarayanan, Inan,
  Shojanazeri, Zou, Wang, Zha, Habeeb, Rudolph, Suk, Aspegren, Goldman, Zhan,
  Damlaj, Molybog, Tufanov, Leontiadis, Veliche, Gat, Weissman, Geboski, Kohli,
  Lam, Asher, Gaya, Marcus, Tang, Chan, Zhen, Reizenstein, Teboul, Zhong, Jin,
  Yang, Cummings, Carvill, Shepard, McPhie, Torres, Ginsburg, Wang, Wu, U,
  Saxena, Khandelwal, Zand, Matosich, Veeraraghavan, Michelena, Li, Jagadeesh,
  Huang, Chawla, Huang, Chen, Garg, A, Silva, Bell, Zhang, Guo, Yu, Moshkovich,
  Wehrstedt, Khabsa, Avalani, Bhatt, Mankus, Hasson, Lennie, Reso, Groshev,
  Naumov, Lathi, Keneally, Liu, Seltzer, Valko, Restrepo, Patel, Vyatskov,
  Samvelyan, Clark, Macey, Wang, Hermoso, Metanat, Rastegari, Bansal,
  Santhanam, Parks, White, Bawa, Singhal, Egebo, Usunier, Mehta, Laptev, Dong,
  Cheng, Chernoguz, Hart, Salpekar, Kalinli, Kent, Parekh, Saab, Balaji,
  Rittner, Bontrager, Roux, Dollar, Zvyagina, Ratanchandani, Yuvraj, Liang,
  Alao, Rodriguez, Ayub, Murthy, Nayani, Mitra, Parthasarathy, Li, Hogan,
  Battey, Wang, Howes, Rinott, Mehta, Siby, Bondu, Datta, Chugh, Hunt, Dhillon,
  Sidorov, Pan, Mahajan, Verma, Yamamoto, Ramaswamy, Lindsay, Lindsay, Feng,
  Lin, Zha, Patil, Shankar, Zhang, Zhang, Wang, Agarwal, Sajuyigbe, Chintala,
  Max, Chen, Kehoe, Satterfield, Govindaprasad, Gupta, Deng, Cho, Virk,
  Subramanian, Choudhury, Goldman, Remez, Glaser, Best, Koehler, Robinson, Li,
  Zhang, Matthews, Chou, Shaked, Vontimitta, Ajayi, Montanez, Mohan, Kumar,
  Mangla, Ionescu, Poenaru, Mihailescu, Ivanov, Li, Wang, Jiang, Bouaziz,
  Constable, Tang, Wu, Wang, Wu, Gao, Kleinman, Chen, Hu, Jia, Qi, Li, Zhang,
  Zhang, Adi, Nam, Yu, Wang, Zhao, Hao, Qian, Li, He, Rait, DeVito, Rosnbrick,
  Wen, Yang, Zhao, and Ma}]{grattafioriLlama3Herd2024}
Aaron Grattafiori, Abhimanyu Dubey, Abhinav Jauhri, Abhinav Pandey, Abhishek
  Kadian, Ahmad {Al-Dahle}, Aiesha Letman, Akhil Mathur, Alan Schelten, Alex
  Vaughan, Amy Yang, Angela Fan, Anirudh Goyal, Anthony Hartshorn, Aobo Yang,
  Archi Mitra, Archie Sravankumar, Artem Korenev, Arthur Hinsvark, Arun Rao,
  Aston Zhang, Aurelien Rodriguez, Austen Gregerson, Ava Spataru, Baptiste
  Roziere, Bethany Biron, Binh Tang, Bobbie Chern, Charlotte Caucheteux, Chaya
  Nayak, Chloe Bi, Chris Marra, Chris McConnell, Christian Keller, Christophe
  Touret, Chunyang Wu, Corinne Wong, Cristian~Canton Ferrer, Cyrus Nikolaidis,
  Damien Allonsius, Daniel Song, Danielle Pintz, Danny Livshits, Danny Wyatt,
  David Esiobu, Dhruv Choudhary, Dhruv Mahajan, Diego {Garcia-Olano}, Diego
  Perino, Dieuwke Hupkes, Egor Lakomkin, Ehab AlBadawy, Elina Lobanova, Emily
  Dinan, Eric~Michael Smith, Filip Radenovic, Francisco Guzm{\'a}n, Frank
  Zhang, Gabriel Synnaeve, Gabrielle Lee, Georgia~Lewis Anderson, Govind
  Thattai, Graeme Nail, Gregoire Mialon, Guan Pang, Guillem Cucurell, Hailey
  Nguyen, Hannah Korevaar, Hu~Xu, Hugo Touvron, Iliyan Zarov, Imanol~Arrieta
  Ibarra, Isabel Kloumann, Ishan Misra, Ivan Evtimov, Jack Zhang, Jade Copet,
  Jaewon Lee, Jan Geffert, Jana Vranes, Jason Park, Jay Mahadeokar, Jeet Shah,
  Jelmer van~der Linde, Jennifer Billock, Jenny Hong, Jenya Lee, Jeremy Fu,
  Jianfeng Chi, Jianyu Huang, Jiawen Liu, Jie Wang, Jiecao Yu, Joanna Bitton,
  Joe Spisak, Jongsoo Park, Joseph Rocca, Joshua Johnstun, Joshua Saxe, Junteng
  Jia, Kalyan~Vasuden Alwala, Karthik Prasad, Kartikeya Upasani, Kate Plawiak,
  Ke~Li, Kenneth Heafield, Kevin Stone, Khalid {El-Arini}, Krithika Iyer,
  Kshitiz Malik, Kuenley Chiu, Kunal Bhalla, Kushal Lakhotia, Lauren
  {Rantala-Yeary}, Laurens van~der Maaten, Lawrence Chen, Liang Tan, Liz
  Jenkins, Louis Martin, Lovish Madaan, Lubo Malo, Lukas Blecher, Lukas
  Landzaat, Luke de~Oliveira, Madeline Muzzi, Mahesh Pasupuleti, Mannat Singh,
  Manohar Paluri, Marcin Kardas, Maria Tsimpoukelli, Mathew Oldham, Mathieu
  Rita, Maya Pavlova, Melanie Kambadur, Mike Lewis, Min Si, Mitesh~Kumar Singh,
  Mona Hassan, Naman Goyal, Narjes Torabi, Nikolay Bashlykov, Nikolay
  Bogoychev, Niladri Chatterji, Ning Zhang, Olivier Duchenne, Onur {\c C}elebi,
  Patrick Alrassy, Pengchuan Zhang, Pengwei Li, Petar Vasic, Peter Weng,
  Prajjwal Bhargava, Pratik Dubal, Praveen Krishnan, Punit~Singh Koura, Puxin
  Xu, Qing He, Qingxiao Dong, Ragavan Srinivasan, Raj Ganapathy, Ramon
  Calderer, Ricardo~Silveira Cabral, Robert Stojnic, Roberta Raileanu, Rohan
  Maheswari, Rohit Girdhar, Rohit Patel, Romain Sauvestre, Ronnie Polidoro,
  Roshan Sumbaly, Ross Taylor, Ruan Silva, Rui Hou, Rui Wang, Saghar Hosseini,
  Sahana Chennabasappa, Sanjay Singh, Sean Bell, Seohyun~Sonia Kim, Sergey
  Edunov, Shaoliang Nie, Sharan Narang, Sharath Raparthy, Sheng Shen, Shengye
  Wan, Shruti Bhosale, Shun Zhang, Simon Vandenhende, Soumya Batra, Spencer
  Whitman, Sten Sootla, Stephane Collot, Suchin Gururangan, Sydney Borodinsky,
  Tamar Herman, Tara Fowler, Tarek Sheasha, Thomas Georgiou, Thomas Scialom,
  Tobias Speckbacher, Todor Mihaylov, Tong Xiao, Ujjwal Karn, Vedanuj Goswami,
  Vibhor Gupta, Vignesh Ramanathan, Viktor Kerkez, Vincent Gonguet, Virginie
  Do, Vish Vogeti, V{\'i}tor Albiero, Vladan Petrovic, Weiwei Chu, Wenhan
  Xiong, Wenyin Fu, Whitney Meers, Xavier Martinet, Xiaodong Wang, Xiaofang
  Wang, Xiaoqing~Ellen Tan, Xide Xia, Xinfeng Xie, Xuchao Jia, Xuewei Wang,
  Yaelle Goldschlag, Yashesh Gaur, Yasmine Babaei, Yi~Wen, Yiwen Song, Yuchen
  Zhang, Yue Li, Yuning Mao, Zacharie~Delpierre Coudert, Zheng Yan, Zhengxing
  Chen, Zoe Papakipos, Aaditya Singh, Aayushi Srivastava, Abha Jain, Adam
  Kelsey, Adam Shajnfeld, Adithya Gangidi, Adolfo Victoria, Ahuva Goldstand,
  Ajay Menon, Ajay Sharma, Alex Boesenberg, Alexei Baevski, Allie Feinstein,
  Amanda Kallet, Amit Sangani, Amos Teo, Anam Yunus, Andrei Lupu, Andres
  Alvarado, Andrew Caples, Andrew Gu, Andrew Ho, Andrew Poulton, Andrew Ryan,
  Ankit Ramchandani, Annie Dong, Annie Franco, Anuj Goyal, Aparajita Saraf,
  Arkabandhu Chowdhury, Ashley Gabriel, Ashwin Bharambe, Assaf Eisenman, Azadeh
  Yazdan, Beau James, Ben Maurer, Benjamin Leonhardi, Bernie Huang, Beth Loyd,
  Beto~De Paola, Bhargavi Paranjape, Bing Liu, Bo~Wu, Boyu Ni, Braden Hancock,
  Bram Wasti, Brandon Spence, Brani Stojkovic, Brian Gamido, Britt Montalvo,
  Carl Parker, Carly Burton, Catalina Mejia, Ce~Liu, Changhan Wang, Changkyu
  Kim, Chao Zhou, Chester Hu, Ching-Hsiang Chu, Chris Cai, Chris Tindal,
  Christoph Feichtenhofer, Cynthia Gao, Damon Civin, Dana Beaty, Daniel
  Kreymer, Daniel Li, David Adkins, David Xu, Davide Testuggine, Delia David,
  Devi Parikh, Diana Liskovich, Didem Foss, Dingkang Wang, Duc Le, Dustin
  Holland, Edward Dowling, Eissa Jamil, Elaine Montgomery, Eleonora Presani,
  Emily Hahn, Emily Wood, Eric-Tuan Le, Erik Brinkman, Esteban Arcaute, Evan
  Dunbar, Evan Smothers, Fei Sun, Felix Kreuk, Feng Tian, Filippos Kokkinos,
  Firat Ozgenel, Francesco Caggioni, Frank Kanayet, Frank Seide,
  Gabriela~Medina Florez, Gabriella Schwarz, Gada Badeer, Georgia Swee, Gil
  Halpern, Grant Herman, Grigory Sizov, Guangyi, Zhang, Guna Lakshminarayanan,
  Hakan Inan, Hamid Shojanazeri, Han Zou, Hannah Wang, Hanwen Zha, Haroun
  Habeeb, Harrison Rudolph, Helen Suk, Henry Aspegren, Hunter Goldman, Hongyuan
  Zhan, Ibrahim Damlaj, Igor Molybog, Igor Tufanov, Ilias Leontiadis,
  Irina-Elena Veliche, Itai Gat, Jake Weissman, James Geboski, James Kohli,
  Janice Lam, Japhet Asher, Jean-Baptiste Gaya, Jeff Marcus, Jeff Tang,
  Jennifer Chan, Jenny Zhen, Jeremy Reizenstein, Jeremy Teboul, Jessica Zhong,
  Jian Jin, Jingyi Yang, Joe Cummings, Jon Carvill, Jon Shepard, Jonathan
  McPhie, Jonathan Torres, Josh Ginsburg, Junjie Wang, Kai Wu, Kam~Hou U, Karan
  Saxena, Kartikay Khandelwal, Katayoun Zand, Kathy Matosich, Kaushik
  Veeraraghavan, Kelly Michelena, Keqian Li, Kiran Jagadeesh, Kun Huang, Kunal
  Chawla, Kyle Huang, Lailin Chen, Lakshya Garg, Lavender A, Leandro Silva, Lee
  Bell, Lei Zhang, Liangpeng Guo, Licheng Yu, Liron Moshkovich, Luca Wehrstedt,
  Madian Khabsa, Manav Avalani, Manish Bhatt, Martynas Mankus, Matan Hasson,
  Matthew Lennie, Matthias Reso, Maxim Groshev, Maxim Naumov, Maya Lathi,
  Meghan Keneally, Miao Liu, Michael~L. Seltzer, Michal Valko, Michelle
  Restrepo, Mihir Patel, Mik Vyatskov, Mikayel Samvelyan, Mike Clark, Mike
  Macey, Mike Wang, Miquel~Jubert Hermoso, Mo~Metanat, Mohammad Rastegari,
  Munish Bansal, Nandhini Santhanam, Natascha Parks, Natasha White, Navyata
  Bawa, Nayan Singhal, Nick Egebo, Nicolas Usunier, Nikhil Mehta,
  Nikolay~Pavlovich Laptev, Ning Dong, Norman Cheng, Oleg Chernoguz, Olivia
  Hart, Omkar Salpekar, Ozlem Kalinli, Parkin Kent, Parth Parekh, Paul Saab,
  Pavan Balaji, Pedro Rittner, Philip Bontrager, Pierre Roux, Piotr Dollar,
  Polina Zvyagina, Prashant Ratanchandani, Pritish Yuvraj, Qian Liang, Rachad
  Alao, Rachel Rodriguez, Rafi Ayub, Raghotham Murthy, Raghu Nayani, Rahul
  Mitra, Rangaprabhu Parthasarathy, Raymond Li, Rebekkah Hogan, Robin Battey,
  Rocky Wang, Russ Howes, Ruty Rinott, Sachin Mehta, Sachin Siby, Sai~Jayesh
  Bondu, Samyak Datta, Sara Chugh, Sara Hunt, Sargun Dhillon, Sasha Sidorov,
  Satadru Pan, Saurabh Mahajan, Saurabh Verma, Seiji Yamamoto, Sharadh
  Ramaswamy, Shaun Lindsay, Shaun Lindsay, Sheng Feng, Shenghao Lin,
  Shengxin~Cindy Zha, Shishir Patil, Shiva Shankar, Shuqiang Zhang, Shuqiang
  Zhang, Sinong Wang, Sneha Agarwal, Soji Sajuyigbe, Soumith Chintala,
  Stephanie Max, Stephen Chen, Steve Kehoe, Steve Satterfield, Sudarshan
  Govindaprasad, Sumit Gupta, Summer Deng, Sungmin Cho, Sunny Virk, Suraj
  Subramanian, Sy~Choudhury, Sydney Goldman, Tal Remez, Tamar Glaser, Tamara
  Best, Thilo Koehler, Thomas Robinson, Tianhe Li, Tianjun Zhang, Tim Matthews,
  Timothy Chou, Tzook Shaked, Varun Vontimitta, Victoria Ajayi, Victoria
  Montanez, Vijai Mohan, Vinay~Satish Kumar, Vishal Mangla, Vlad Ionescu, Vlad
  Poenaru, Vlad~Tiberiu Mihailescu, Vladimir Ivanov, Wei Li, Wenchen Wang,
  Wenwen Jiang, Wes Bouaziz, Will Constable, Xiaocheng Tang, Xiaojian Wu,
  Xiaolan Wang, Xilun Wu, Xinbo Gao, Yaniv Kleinman, Yanjun Chen, Ye~Hu,
  Ye~Jia, Ye~Qi, Yenda Li, Yilin Zhang, Ying Zhang, Yossi Adi, Youngjin Nam,
  Yu, Wang, Yu~Zhao, Yuchen Hao, Yundi Qian, Yunlu Li, Yuzi He, Zach Rait,
  Zachary DeVito, Zef Rosnbrick, Zhaoduo Wen, Zhenyu Yang, Zhiwei Zhao, and
  Zhiyu Ma. 2024.
\newblock \href {https://doi.org/10.48550/arXiv.2407.21783} {The {{Llama}} 3
  {{Herd}} of {{Models}}}.
\newblock \emph{Preprint}, arXiv:2407.21783.

\bibitem[{Joshi et~al.(2017)Joshi, Choi, Weld, and
  Zettlemoyer}]{joshi-etal-2017-triviaqa}
Mandar Joshi, Eunsol Choi, Daniel Weld, and Luke Zettlemoyer. 2017.
\newblock \href {https://doi.org/10.18653/v1/P17-1147} {{T}rivia{QA}: A large
  scale distantly supervised challenge dataset for reading comprehension}.
\newblock In \emph{Proceedings of the 55th Annual Meeting of the Association
  for Computational Linguistics (Volume 1: Long Papers)}, pages 1601--1611,
  Vancouver, Canada. Association for Computational Linguistics.

\bibitem[{Kojima et~al.(2022)Kojima, Gu, Reid, Matsuo, and
  Iwasawa}]{kojimaLargeLanguageModels2022}
Takeshi Kojima, S.~Gu, Machel Reid, Yutaka Matsuo, and Yusuke Iwasawa. 2022.
\newblock \href
  {https://www.semanticscholar.org/paper/Large-Language-Models-are-Zero-Shot-Reasoners-Kojima-Gu/e7ad08848d5d7c5c47673ffe0da06af443643bda}
  {Large {{Language Models}} are {{Zero-Shot Reasoners}}}.
\newblock \emph{ArXiv}.

\bibitem[{OpenAI et~al.(2024)OpenAI, Hurst, Lerer, Goucher, Perelman, Ramesh,
  Clark, Ostrow, Welihinda, Hayes, Radford, M{\k a}dry, {Baker-Whitcomb},
  Beutel, Borzunov, Carney, Chow, Kirillov, Nichol, Paino, Renzin, Passos,
  Kirillov, Christakis, Conneau, Kamali, Jabri, Moyer, Tam, Crookes,
  Tootoochian, Tootoonchian, Kumar, Vallone, Karpathy, Braunstein, Cann,
  Codispoti, Galu, Kondrich, Tulloch, Mishchenko, Baek, Jiang, Pelisse,
  Woodford, Gosalia, Dhar, Pantuliano, Nayak, Oliver, Zoph, Ghorbani,
  Leimberger, Rossen, Sokolowsky, Wang, Zweig, Hoover, Samic, McGrew, Spero,
  Giertler, Cheng, Lightcap, Walkin, Quinn, Guarraci, Hsu, Kellogg, Eastman,
  Lugaresi, Wainwright, Bassin, Hudson, Chu, Nelson, Li, Shern, Conger,
  Barette, Voss, Ding, Lu, Zhang, Beaumont, Hallacy, Koch, Gibson, Kim, Choi,
  McLeavey, Hesse, Fischer, Winter, Czarnecki, Jarvis, Wei, Koumouzelis,
  Sherburn, Kappler, Levin, Levy, Carr, Farhi, Mely, Robinson, Sasaki, Jin,
  Valladares, Tsipras, Li, Nguyen, Findlay, Oiwoh, Wong, Asdar, Proehl, Yang,
  Antonow, Kramer, Peterson, Sigler, Wallace, Brevdo, Mays, Khorasani, Such,
  Raso, Zhang, von Lohmann, Sulit, Goh, Oden, Salmon, Starace, Brockman,
  Salman, Bao, Hu, Wong, Wang, Schmidt, Whitney, Jun, Kirchner, Pinto, Ren,
  Chang, Chung, Kivlichan, O'Connell, O'Connell, Osband, Silber, Sohl, Okuyucu,
  Lan, Kostrikov, Sutskever, Kanitscheider, Gulrajani, Coxon, Menick, Pachocki,
  Aung, Betker, Crooks, Lennon, Kiros, Leike, Park, Kwon, Phang, Teplitz, Wei,
  Wolfe, Chen, Harris, Varavva, Lee, Shieh, Lin, Yu, Weng, Tang, Yu, Jang,
  Candela, Beutler, Landers, Parish, Heidecke, Schulman, Lachman, McKay,
  Uesato, Ward, Kim, Huizinga, Sitkin, Kraaijeveld, Gross, Kaplan, Snyder,
  Achiam, Jiao, Lee, Zhuang, Harriman, Fricke, Hayashi, Singhal, Shi, Karthik,
  Wood, Rimbach, Hsu, Nguyen, {Gu-Lemberg}, Button, Liu, Howe, Muthukumar,
  Luther, Ahmad, Kai, Itow, Workman, Pathak, Chen, Jing, Guy, Fedus, Zhou,
  Mamitsuka, Weng, McCallum, Held, Ouyang, Feuvrier, Zhang, Kondraciuk, Kaiser,
  Hewitt, Metz, Doshi, Aflak, Simens, Boyd, Thompson, Dukhan, Chen, Gray,
  Hudnall, Zhang, Aljubeh, Litwin, Zeng, Johnson, Shetty, Gupta, Shah, Yatbaz,
  Yang, Zhong, Glaese, Chen, Janner, Lampe, Petrov, Wu, Wang, Fradin, Pokrass,
  Castro, de~Castro, Pavlov, Brundage, Wang, Khan, Murati, Bavarian, Lin,
  Yesildal, Soto, Gimelshein, Cone, Staudacher, Summers, LaFontaine, Chowdhury,
  Ryder, Stathas, Turley, Tezak, Felix, Kudige, Keskar, Deutsch, Bundick,
  Puckett, Nachum, Okelola, Boiko, Murk, Jaffe, Watkins, Godement,
  {Campbell-Moore}, Chao, McMillan, Belov, Su, Bak, Bakkum, Deng, Dolan,
  Hoeschele, Welinder, Tillet, Pronin, Tillet, Dhariwal, Yuan, Dias, Lim,
  Arora, Troll, Lin, Lopes, Puri, Miyara, Leike, Gaubert, Zamani, Wang,
  Donnelly, Honsby, Smith, Sahai, Ramchandani, Huet, Carmichael, Zellers, Chen,
  Chen, Nigmatullin, Cheu, Jain, Altman, Schoenholz, Toizer, Miserendino,
  Agarwal, Culver, Ethersmith, Gray, Grove, Metzger, Hermani, Jain, Zhao, Wu,
  Jomoto, Wu, Shuaiqi, Xia, Phene, Papay, Narayanan, Coffey, Lee, Hall, Balaji,
  Broda, Stramer, Xu, Gogineni, Christianson, Sanders, Patwardhan, Cunninghman,
  Degry, Dimson, Raoux, Shadwell, Zheng, Underwood, Markov, Sherbakov, Rubin,
  Stasi, Kaftan, Heywood, Peterson, Walters, Eloundou, Qi, Moeller, Monaco,
  Kuo, Fomenko, Chang, Zheng, Zhou, Manassra, Sheu, Zaremba, Patil, Qian, Kim,
  Cheng, Zhang, He, Zhang, Jin, Dai, and Malkov}]{openaiGPT4oSystemCard2024}
OpenAI, Aaron Hurst, Adam Lerer, Adam~P. Goucher, Adam Perelman, Aditya Ramesh,
  Aidan Clark, A.~J. Ostrow, Akila Welihinda, Alan Hayes, Alec Radford,
  Aleksander M{\k a}dry, Alex {Baker-Whitcomb}, Alex Beutel, Alex Borzunov,
  Alex Carney, Alex Chow, Alex Kirillov, Alex Nichol, Alex Paino, Alex Renzin,
  Alex~Tachard Passos, Alexander Kirillov, Alexi Christakis, Alexis Conneau,
  Ali Kamali, Allan Jabri, Allison Moyer, Allison Tam, Amadou Crookes, Amin
  Tootoochian, Amin Tootoonchian, Ananya Kumar, Andrea Vallone, Andrej
  Karpathy, Andrew Braunstein, Andrew Cann, Andrew Codispoti, Andrew Galu,
  Andrew Kondrich, Andrew Tulloch, Andrey Mishchenko, Angela Baek, Angela
  Jiang, Antoine Pelisse, Antonia Woodford, Anuj Gosalia, Arka Dhar, Ashley
  Pantuliano, Avi Nayak, Avital Oliver, Barret Zoph, Behrooz Ghorbani, Ben
  Leimberger, Ben Rossen, Ben Sokolowsky, Ben Wang, Benjamin Zweig, Beth
  Hoover, Blake Samic, Bob McGrew, Bobby Spero, Bogo Giertler, Bowen Cheng,
  Brad Lightcap, Brandon Walkin, Brendan Quinn, Brian Guarraci, Brian Hsu,
  Bright Kellogg, Brydon Eastman, Camillo Lugaresi, Carroll Wainwright, Cary
  Bassin, Cary Hudson, Casey Chu, Chad Nelson, Chak Li, Chan~Jun Shern,
  Channing Conger, Charlotte Barette, Chelsea Voss, Chen Ding, Cheng Lu, Chong
  Zhang, Chris Beaumont, Chris Hallacy, Chris Koch, Christian Gibson, Christina
  Kim, Christine Choi, Christine McLeavey, Christopher Hesse, Claudia Fischer,
  Clemens Winter, Coley Czarnecki, Colin Jarvis, Colin Wei, Constantin
  Koumouzelis, Dane Sherburn, Daniel Kappler, Daniel Levin, Daniel Levy, David
  Carr, David Farhi, David Mely, David Robinson, David Sasaki, Denny Jin, Dev
  Valladares, Dimitris Tsipras, Doug Li, Duc~Phong Nguyen, Duncan Findlay,
  Edede Oiwoh, Edmund Wong, Ehsan Asdar, Elizabeth Proehl, Elizabeth Yang, Eric
  Antonow, Eric Kramer, Eric Peterson, Eric Sigler, Eric Wallace, Eugene
  Brevdo, Evan Mays, Farzad Khorasani, Felipe~Petroski Such, Filippo Raso,
  Francis Zhang, Fred von Lohmann, Freddie Sulit, Gabriel Goh, Gene Oden, Geoff
  Salmon, Giulio Starace, Greg Brockman, Hadi Salman, Haiming Bao, Haitang Hu,
  Hannah Wong, Haoyu Wang, Heather Schmidt, Heather Whitney, Heewoo Jun,
  Hendrik Kirchner, Henrique Ponde de~Oliveira Pinto, Hongyu Ren, Huiwen Chang,
  Hyung~Won Chung, Ian Kivlichan, Ian O'Connell, Ian O'Connell, Ian Osband, Ian
  Silber, Ian Sohl, Ibrahim Okuyucu, Ikai Lan, Ilya Kostrikov, Ilya Sutskever,
  Ingmar Kanitscheider, Ishaan Gulrajani, Jacob Coxon, Jacob Menick, Jakub
  Pachocki, James Aung, James Betker, James Crooks, James Lennon, Jamie Kiros,
  Jan Leike, Jane Park, Jason Kwon, Jason Phang, Jason Teplitz, Jason Wei,
  Jason Wolfe, Jay Chen, Jeff Harris, Jenia Varavva, Jessica~Gan Lee, Jessica
  Shieh, Ji~Lin, Jiahui Yu, Jiayi Weng, Jie Tang, Jieqi Yu, Joanne Jang,
  Joaquin~Quinonero Candela, Joe Beutler, Joe Landers, Joel Parish, Johannes
  Heidecke, John Schulman, Jonathan Lachman, Jonathan McKay, Jonathan Uesato,
  Jonathan Ward, Jong~Wook Kim, Joost Huizinga, Jordan Sitkin, Jos Kraaijeveld,
  Josh Gross, Josh Kaplan, Josh Snyder, Joshua Achiam, Joy Jiao, Joyce Lee,
  Juntang Zhuang, Justyn Harriman, Kai Fricke, Kai Hayashi, Karan Singhal, Katy
  Shi, Kavin Karthik, Kayla Wood, Kendra Rimbach, Kenny Hsu, Kenny Nguyen,
  Keren {Gu-Lemberg}, Kevin Button, Kevin Liu, Kiel Howe, Krithika Muthukumar,
  Kyle Luther, Lama Ahmad, Larry Kai, Lauren Itow, Lauren Workman, Leher
  Pathak, Leo Chen, Li~Jing, Lia Guy, Liam Fedus, Liang Zhou, Lien Mamitsuka,
  Lilian Weng, Lindsay McCallum, Lindsey Held, Long Ouyang, Louis Feuvrier,
  Lu~Zhang, Lukas Kondraciuk, Lukasz Kaiser, Luke Hewitt, Luke Metz, Lyric
  Doshi, Mada Aflak, Maddie Simens, Madelaine Boyd, Madeleine Thompson, Marat
  Dukhan, Mark Chen, Mark Gray, Mark Hudnall, Marvin Zhang, Marwan Aljubeh,
  Mateusz Litwin, Matthew Zeng, Max Johnson, Maya Shetty, Mayank Gupta, Meghan
  Shah, Mehmet Yatbaz, Meng~Jia Yang, Mengchao Zhong, Mia Glaese, Mianna Chen,
  Michael Janner, Michael Lampe, Michael Petrov, Michael Wu, Michele Wang,
  Michelle Fradin, Michelle Pokrass, Miguel Castro, Miguel Oom~Temudo
  de~Castro, Mikhail Pavlov, Miles Brundage, Miles Wang, Minal Khan, Mira
  Murati, Mo~Bavarian, Molly Lin, Murat Yesildal, Nacho Soto, Natalia
  Gimelshein, Natalie Cone, Natalie Staudacher, Natalie Summers, Natan
  LaFontaine, Neil Chowdhury, Nick Ryder, Nick Stathas, Nick Turley, Nik Tezak,
  Niko Felix, Nithanth Kudige, Nitish Keskar, Noah Deutsch, Noel Bundick, Nora
  Puckett, Ofir Nachum, Ola Okelola, Oleg Boiko, Oleg Murk, Oliver Jaffe,
  Olivia Watkins, Olivier Godement, Owen {Campbell-Moore}, Patrick Chao, Paul
  McMillan, Pavel Belov, Peng Su, Peter Bak, Peter Bakkum, Peter Deng, Peter
  Dolan, Peter Hoeschele, Peter Welinder, Phil Tillet, Philip Pronin, Philippe
  Tillet, Prafulla Dhariwal, Qiming Yuan, Rachel Dias, Rachel Lim, Rahul Arora,
  Rajan Troll, Randall Lin, Rapha~Gontijo Lopes, Raul Puri, Reah Miyara, Reimar
  Leike, Renaud Gaubert, Reza Zamani, Ricky Wang, Rob Donnelly, Rob Honsby,
  Rocky Smith, Rohan Sahai, Rohit Ramchandani, Romain Huet, Rory Carmichael,
  Rowan Zellers, Roy Chen, Ruby Chen, Ruslan Nigmatullin, Ryan Cheu, Saachi
  Jain, Sam Altman, Sam Schoenholz, Sam Toizer, Samuel Miserendino, Sandhini
  Agarwal, Sara Culver, Scott Ethersmith, Scott Gray, Sean Grove, Sean Metzger,
  Shamez Hermani, Shantanu Jain, Shengjia Zhao, Sherwin Wu, Shino Jomoto,
  Shirong Wu, Shuaiqi, Xia, Sonia Phene, Spencer Papay, Srinivas Narayanan,
  Steve Coffey, Steve Lee, Stewart Hall, Suchir Balaji, Tal Broda, Tal Stramer,
  Tao Xu, Tarun Gogineni, Taya Christianson, Ted Sanders, Tejal Patwardhan,
  Thomas Cunninghman, Thomas Degry, Thomas Dimson, Thomas Raoux, Thomas
  Shadwell, Tianhao Zheng, Todd Underwood, Todor Markov, Toki Sherbakov, Tom
  Rubin, Tom Stasi, Tomer Kaftan, Tristan Heywood, Troy Peterson, Tyce Walters,
  Tyna Eloundou, Valerie Qi, Veit Moeller, Vinnie Monaco, Vishal Kuo, Vlad
  Fomenko, Wayne Chang, Weiyi Zheng, Wenda Zhou, Wesam Manassra, Will Sheu,
  Wojciech Zaremba, Yash Patil, Yilei Qian, Yongjik Kim, Youlong Cheng,
  Yu~Zhang, Yuchen He, Yuchen Zhang, Yujia Jin, Yunxing Dai, and Yury Malkov.
  2024.
\newblock \href {https://doi.org/10.48550/arXiv.2410.21276} {{{GPT-4o System
  Card}}}.
\newblock \emph{Preprint}, arXiv:2410.21276.

\bibitem[{Ouyang et~al.(2022)Ouyang, Wu, Jiang, Almeida, Wainwright, Mishkin,
  Zhang, Agarwal, Slama, Ray, Schulman, Hilton, Kelton, Miller, Simens, Askell,
  Welinder, Christiano, Leike, and Lowe}]{ouyangTrainingLanguageModels2022}
Long Ouyang, Jeff Wu, Xu~Jiang, Diogo Almeida, Carroll~L. Wainwright, Pamela
  Mishkin, Chong Zhang, Sandhini Agarwal, Katarina Slama, Alex Ray, John
  Schulman, Jacob Hilton, Fraser Kelton, Luke Miller, Maddie Simens, Amanda
  Askell, Peter Welinder, Paul Christiano, Jan Leike, and Ryan Lowe. 2022.
\newblock \href {https://doi.org/10.48550/arXiv.2203.02155} {Training language
  models to follow instructions with human feedback}.
\newblock \emph{Preprint}, arXiv:2203.02155.

\bibitem[{Snell et~al.(2024)Snell, Lee, Xu, and
  Kumar}]{snellScalingLLMTestTime2024}
Charlie Snell, Jaehoon Lee, Kelvin Xu, and Aviral Kumar. 2024.
\newblock \href {https://doi.org/10.48550/arXiv.2408.03314} {Scaling {{LLM
  Test-Time Compute Optimally}} can be {{More Effective}} than {{Scaling Model
  Parameters}}}.
\newblock \emph{Preprint}, arXiv:2408.03314.

\bibitem[{Stelmakh et~al.(2022)Stelmakh, Luan, Dhingra, and
  Chang}]{stelmakh-etal-2022-asqa}
Ivan Stelmakh, Yi~Luan, Bhuwan Dhingra, and Ming-Wei Chang. 2022.
\newblock \href {https://doi.org/10.18653/v1/2022.emnlp-main.566} {{ASQA}:
  Factoid questions meet long-form answers}.
\newblock In \emph{Proceedings of the 2022 Conference on Empirical Methods in
  Natural Language Processing}, pages 8273--8288, Abu Dhabi, United Arab
  Emirates. Association for Computational Linguistics.

\bibitem[{Vaswani et~al.(2017)Vaswani, Shazeer, Parmar, Uszkoreit, Jones,
  Gomez, Kaiser, and Polosukhin}]{vaswaniAttentionAllYou2017}
Ashish Vaswani, Noam Shazeer, Niki Parmar, Jakob Uszkoreit, Llion Jones,
  Aidan~N. Gomez, Lukasz Kaiser, and Illia Polosukhin. 2017.
\newblock \href {https://doi.org/10.48550/arXiv.1706.03762} {Attention {{Is All
  You Need}}}.
\newblock \emph{Preprint}, arXiv:1706.03762.

\bibitem[{Wang et~al.(2023)Wang, Wei, Schuurmans, Le, Chi, Narang, Chowdhery,
  and Zhou}]{wangSelfConsistencyImprovesChain2023}
Xuezhi Wang, Jason Wei, Dale Schuurmans, Quoc Le, Ed~Chi, Sharan Narang,
  Aakanksha Chowdhery, and Denny Zhou. 2023.
\newblock \href {https://arxiv.org/abs/2203.11171} {Self-{{Consistency Improves
  Chain}} of {{Thought Reasoning}} in {{Language Models}}}.
\newblock \emph{Preprint}, arXiv:2203.11171.

\bibitem[{Wei et~al.(2022)Wei, Bosma, Zhao, Guu, Yu, Lester, Du, Dai, and
  Le}]{weiFinetunedLanguageModels2022}
Jason Wei, Maarten Bosma, Vincent~Y. Zhao, Kelvin Guu, Adams~Wei Yu, Brian
  Lester, Nan Du, Andrew~M. Dai, and Quoc~V. Le. 2022.
\newblock \href {https://doi.org/10.48550/arXiv.2109.01652} {Finetuned
  {{Language Models Are Zero-Shot Learners}}}.
\newblock \emph{Preprint}, arXiv:2109.01652.

\bibitem[{Wei et~al.(2023)Wei, Wang, Schuurmans, Bosma, Ichter, Xia, Chi, Le,
  and Zhou}]{weiChainofThoughtPromptingElicits2023}
Jason Wei, Xuezhi Wang, Dale Schuurmans, Maarten Bosma, Brian Ichter, Fei Xia,
  Ed~Chi, Quoc Le, and Denny Zhou. 2023.
\newblock \href {https://doi.org/10.48550/arXiv.2201.11903} {Chain-of-{{Thought
  Prompting Elicits Reasoning}} in {{Large Language Models}}}.
\newblock \emph{Preprint}, arXiv:2201.11903.

\bibitem[{Wu et~al.(2023)Wu, Zhang, and Huang}]{wu-etal-2023-chain}
Dingjun Wu, Jing Zhang, and Xinmei Huang. 2023.
\newblock \href {https://doi.org/10.18653/v1/2023.findings-acl.408} {Chain of
  thought prompting elicits knowledge augmentation}.
\newblock In \emph{Findings of the Association for Computational Linguistics:
  ACL 2023}, pages 6519--6534, Toronto, Canada. Association for Computational
  Linguistics.

\bibitem[{Yang et~al.(2018)Yang, Qi, Zhang, Bengio, Cohen, Salakhutdinov, and
  Manning}]{yang-etal-2018-hotpotqa}
Zhilin Yang, Peng Qi, Saizheng Zhang, Yoshua Bengio, William Cohen, Ruslan
  Salakhutdinov, and Christopher~D. Manning. 2018.
\newblock \href {https://doi.org/10.18653/v1/D18-1259} {{H}otpot{QA}: A dataset
  for diverse, explainable multi-hop question answering}.
\newblock In \emph{Proceedings of the 2018 Conference on Empirical Methods in
  Natural Language Processing}, pages 2369--2380, Brussels, Belgium.
  Association for Computational Linguistics.

\bibitem[{Yen et~al.(2024)Yen, Gao, Hou, Ding, Fleischer, Izsak, Wasserblat,
  and Chen}]{yenHELMETHowEvaluate2024}
Howard Yen, Tianyu Gao, Minmin Hou, Ke~Ding, Daniel Fleischer, Peter Izsak,
  Moshe Wasserblat, and Danqi Chen. 2024.
\newblock \href {https://arxiv.org/abs/2410.02694v1} {{{HELMET}}: {{How}} to
  {{Evaluate Long-Context Language Models Effectively}} and {{Thoroughly}}}.

\bibitem[{Zhang et~al.(2023)Zhang, Xiao, Liu, Dou, and
  Nie}]{zhangRetrieveAnythingAugment2023}
Peitian Zhang, Shitao Xiao, Zheng Liu, Zhicheng Dou, and Jian-Yun Nie. 2023.
\newblock \href {https://arxiv.org/abs/2310.07554} {Retrieve {{Anything To
  Augment Large Language Models}}}.
\newblock \emph{Preprint}, arXiv:2310.07554.

\end{thebibliography}

\appendix
\newpage
\clearpage
\section{Implementation Details}

\subsection{Datasets}
\label{sec:datasets}

Datasets used:
\begin{itemize}[noitemsep,topsep=0.5em,parsep=0.4em,leftmargin=1.5em]
\item TriviaQA: {\small \url{https://huggingface.co/datasets/Tevatron/wikipedia-trivia}}
\item HotpotQA: {\small \url{https://huggingface.co/datasets/facebook/kilt_tasks/viewer/hotpotqa}}
\item ASQA: {\small \url{https://huggingface.co/datasets/din0s/asqa}}
\end{itemize}

\noindent
Random sample of size 200 was used with each dataset. External context is comprised of top \mbox{$k=5$} passages. TriviaQA dataset contains external context. For HotpotQA and ASQA, retrieval was done over a Wikipedia corpus based on a December 2021 dump, using the \href{https://huggingface.co/BAAI/llm-embedder}{BAAI/llm-embedder} dense embedder \cite{zhangRetrieveAnythingAugment2023}.

\subsection{Prompts and Examples}
\label{sec:prompts}

\begin{itemize}

\item \rephrase{} system prompt: \ref{tab:square_prompt}.
\item \rephrase{} few-shot examples: \ref{tab:fewshot_example}.

\item \textit{Summarize} system prompt: \ref{tab:summary_prompt}
\item \textit{Summarize} few-shot examples: \ref{tab:fewshot_summary_example}.    

\item \textit{Vote} system prompt: \ref{tab:voting_prompt}.
\item \textit{Vote} few-shot examples: \ref{tab:fewshot_voting_example}.

\item CoT system prompt: \ref{tab:cot_prompt}.
\item CoT few-shot examples: \ref{tab:fewshot_cot_example}.

\item RaR system prompt: \ref{tab:rar_prompt}.
\item RaR few-shot examples: \ref{tab:fewshot_rar_example}.

\end{itemize}

\begin{table*}[t]
    \centering
    \small
    \begin{tabular}{llccccccc}
      \toprule
      Dataset & Model & Baseline & RAG & CoT & RaR & \rephrase-3 & \rephrase-5 & \rephrase-10 \\
      \midrule
      \multirow[l]{3}{*}{TriviaQA}
              & Llama-3.2 3B  & 59.5 & 82.0 & 57.0 / 87.5  & 67.5 / 86.0 &  89.5 / 88.5 & 89.5 / 89.0  &  86.0 / 86.5 \\
              & Llama-3.1 8B  & 76.5 & 89.5 & 53.5 / 90.5  & 84.0 / 89.5 &  72.5 / 92.5 & 89.0 / 94.0  &  76.5 / 94.0 \\
              & GPT-4o    & 88.7 & 92.7 & 01.3 / 92.7  & 87.3 / 94.7 &  84.7 / 96.7 & 84.0 / 96.7   &  86.0 / 95.3 \\
      \midrule
      \multirow[l]{3}{*}{HotpotQA}
              & Llama-3.2 3B  &  17.5  & 26.0  & 23.0 / 26.5  & 25.0 / 25.0  &  31.0 / 31.5  & 28.0 / 29.5  &  30.5 / 28.0   \\
              & Llama-3.1 8B  &  23.0  & 26.5  & 21.0 / 31.0  & 26.0 / 28.5  &  30.5 / 33.5  & 33.5 / 31.5  &  29.5 / 33.5   \\
              & GPT-4o    &  44.0  & 45.3  & 06.7 / 46.7  & 40.0 / 47.3  &  45.3 / 46.7  & 44.7 / 47.3  &  40.0 / 46.0   \\
      \midrule
      \multirow[l]{3}{*}{ASQA}
              & Llama-3.2 3B  &  14.2 & 21.5  & 24.4 / 21.9  & 18.3 / 23.5  &  24.2 / 26.6   & 24.8 / 27.9  &  25.6 / 27.4  \\ 
              & Llama-3.1 8B  &  14.6 & 23.1  & 21.2 / 24.8  & 25.1 / 25.5  &  23.8 / 28.8   & 29.1 / 27.9  &  30.0 / 27.8    \\ 
              & GPT-4o    &  26.8 & 30.4  & 02.1 / 31.9  & 20.7 / 30.1  &  24.8 / 31.7    &  30.2 / 29.4 &  29.6 / 31.1   \\ 
      \bottomrule
    \end{tabular}
    \caption{Summary of results; comparing seven configurations on three datasets, using three models. When applicable, zero-shot/few-shot results are shown.}
    \label{tab:more_results}
\end{table*}

\begin{table*}
    \caption{Generated answer (verbatim) for a question from the HotpotQA dataset, produced by Llama-3.1 8B model, using the CoT technique with two few-shot examples. Correct answer is \textbf{March and April}, which is found in the reasoning chain, probably due to this fact appearing in the relevant external context provided to the model (not shown here). The final model's answer is wrong, necessitating the need to correctly extract answers for better analysis and research.}
    \centering
    \small
    \begin{tabular}{>{\raggedright\arraybackslash\tt}p{0.98\textwidth}<{}}
      \toprule
      Question: In what month is the annual documentary film festival, that is presented by the fortnightly published British journal of literary essays, held? \\
      \midrule
      The annual documentary film festival, presented by the fortnightly published British journal, is the Open City Documentary Festival. \\
      \\
      Let's break down the context: \\
      \\
      The journal is described as a fortnightly published British journal of literary essays. \\
      The London Review of Books is mentioned as a publication associated with the London International Documentary Festival, but it is a monthly publication, not fortnightly. \\
      The Open City Documentary Festival is not mentioned in the context as being associated with a specific journal, but it is a  documentary film festival that takes place in London. \\
      However, the Open City Documentary Festival is not the only festival that matches the description. The London Review of Bookseller is a monthly publication, but the journal described in the question is fortnightly. The London Review of Books is not the journal described in the question. \\
      The Open City Documentary Festival is not the only festival that matches the description, but it is the only festival that is associated with a journal that is published twice a year, which could be interpreted as fortnightly. \\
      However, the London Independent Film Festival is not the correct answer because it is held in April, and the journal is not mentioned in the context. \\
      The London International Documentary Festival is held in \textcolor{blue}{\textbf{March and April}}, but the journal is not mentioned in the context. \\
      The Open City Documentary Festival is the only festival that matches the description of being associated with a journal that is published twice a year, which could be interpreted as fortnightly. \\
      \\
      Answer: \textcolor{red}{\textbf{Open City}}. \\
      \bottomrule
    \end{tabular}
    \label{tab:bad-example}
\end{table*}

\begin{table*}
    \caption{The two examples used in our few-shot configurations. Here, the \rephrase{} prompt is used, with $N=3$.}
    \centering
    \small
    \begin{tabular}{>{\raggedright\arraybackslash\tt}p{0.98\textwidth}<{}}
      \toprule
      Question: What is the shared profession of Jack Kerouac and Dan Masterson? \\
      \midrule
      Generated Questions and Answers: \\
      1. Question: What literary movement was Jack Kerouac a pioneer of? \\
      Answer: Jack Kerouac was a pioneer of the Beat Generation. \\
 \\
      2. Question: What is Dan Masterson primarily known for? \\
      Answer: Dan Masterson is primarily known for being a poet. \\
 \\
      3. Question: When was the Jack Kerouac School of Disembodied Poetics founded and by whom? \\
      Answer: The Jack Kerouac School of Disembodied Poetics was founded in 1974 by Allen Ginsberg and Anne Waldman. \\
 \\
      Answer: The shared profession of Jack Kerouac and Dan Masterson is that they were both poets. \\
      \midrule
      Question: Which band top the UK charts with their singles "Chelsea Dagger" and "Whistle for the Choir": The Fratellis or The Madden Brothers? \\
      \midrule
      Questions and Answers Based on the Given Context: \\
 \\
      1. Question: When was "Whistle for the Choir" released, and how did it perform on the UK Singles Chart? \\
      Answer: "Whistle for the Choir" was released on 27 November 2006 and reached number nine on the UK Singles Chart. \\
 \\
      2. Question: What inspired the name of the song "Chelsea Dagger" by The Fratellis? \\
      Answer: The song "Chelsea Dagger" was named after Jon Fratelli's wife Heather, who performed as a burlesque dancer under the stage name Chelsea Dagger. \\
 \\
      3. Question: What is the relationship between the members of The Fratellis, and what are their stage names? \\
      Answer: The members of The Fratellis are not related by blood. They perform under pseudonyms: Jon Fratelli (lead vocalist and guitarist), Barry Fratelli (bassist), and Mince Fratelli (drummer). \\
 \\
      Answer: The Fratellis are the band that topped the UK charts with their singles "Chelsea Dagger" and "Whistle for the Choir." \\
      \bottomrule
    \end{tabular}
    \label{tab:fewshot_example}
\end{table*}

\clearpage

\begin{table*}
    \caption{Prompt used for the \rephrase{} supplement aggregation method, using answer summarization; see \Cref{sec:ablation-study}.}
    \centering
    \small
    \begin{tabular}{>{\raggedright\arraybackslash\tt}p{0.98\textwidth}<{}}
      \toprule
      You are a helpful question answerer who can provide an answer given a question and relevant context. \\
      Generate \{N\} questions based on the given question and context, and shortly answer them. \\
      Shortly summarize what you have learned from the answers. \\
      Finally, provide an answer to the original question using they summary you created. \\
      The answer should be a short span, just a few words. \\
      \bottomrule
    \end{tabular}
    \label{tab:summary_prompt}
\end{table*}

\begin{table*}
    \caption{Two-shot examples used in an ablation study where \rephrase{} is supplemented by an instruction to summarize the model's learning before providing the final answer, see \Cref{sec:ablation-study}.}
    \centering
    \small
    \begin{tabular}{>{\raggedright\arraybackslash\tt}p{0.98\textwidth}<{}}
      \toprule
      Question: What is the shared profession of Jack Kerouac and Dan Masterson? \\
      \midrule
      1. What notable literary movement was Jack Kerouac a pioneer of? \\
      - The Beat Generation.\\
      \\
      2. What type of school is the Jack Kerouac School of Disembodied Poetics?\\
      - A creative writing and literature school.\\
      \\
      3. What is the connection between Jack Kerouac's novel "On the Road" and his personal life?\\
      - The novel is based on his travels and features characters representing key figures of the Beat movement, including himself.\\
      \\
      Summary:\\
      From the provided context, we learned that Jack Kerouac was an American novelist and poet, recognized as a pioneer of the Beat Generation, and known for his spontaneous prose. Dan Masterson was also an American poet, known for his work and his background. Both individuals are noted primarily for their contributions to poetry.\\
      \\
      Answer: Poet.\\
      \midrule
      Question: Which band top the UK charts with their singles "Chelsea Dagger" and "Whistle for the Choir": The Fratellis or The Madden Brothers? \\
      \midrule
      1. Question: What are the names of the members of The Fratellis? \\
      Answer: Jon Fratelli, Barry Fratelli, and Mince Fratelli.\\
      \\
      2. Question: What is the highest chart position "Whistle for the Choir" reached in Scotland?\\ 
      Answer: Number two.\\
      \\
      3. Question: When was "Chelsea Dagger" released as a single?\\
      Answer: 28 August 2006.\\
      \\
      Summary: The Fratellis are a Scottish rock band consisting of Jon, Barry, and Mince Fratelli. Their singles "Chelsea Dagger" and "Whistle for the Choir" were top ten hits in the UK charts, with "Whistle for the Choir" reaching number nine on the UK Singles Chart and number two in Scotland. \\
      \\
      Answer: The Fratellis.\\
      \bottomrule
    \end{tabular}
    \label{tab:fewshot_summary_example}
\end{table*}

\clearpage

\begin{table*}
    \caption{Prompt used for the \rephrase{} supplement aggregation method, using majority voting; see \Cref{sec:ablation-study}.}
    \centering
    \small
    \begin{tabular}{>{\raggedright\arraybackslash\tt}p{0.98\textwidth}<{}}
      \toprule
      You are a helpful question answerer who can provide an answer given a question and relevant context. \\
      Generate \{N\} questions based on the given question and context, and shortly answer them. \\
      Finally, provide an answer to the original question by choosing amongst the answers you created the most common answer. The answer can't be any other option. \\
      The answer should be a short span, just a few words. \\
      \bottomrule
    \end{tabular}
    \label{tab:voting_prompt}
\end{table*}

\begin{table*}
    \caption{Two-shot examples used in an ablation study where \rephrase{} is supplemented by an instruction to pick the final answer as a majority vote amongst the sub-questions' answers, see \Cref{sec:ablation-study}.}
    \centering
    \small
    \begin{tabular}{>{\raggedright\arraybackslash\tt}p{0.98\textwidth}<{}}
      \toprule
      Question: What is the shared profession of Jack Kerouac and Dan Masterson? \\
      \midrule
      1. What was Jack Kerouac's profession? \\
      - Novelist and poet \\
      \\
      2. What was Dan Masterson's profession?\\
      - Poet\\
      \\
      3. What genre is Jack Kerouac associated with?\\
      - Beat Generation literature\\
      \\
      Answer: Poet\\
      \midrule
      Question: Which band top the UK charts with their singles "Chelsea Dagger" and "Whistle for the Choir": The Fratellis or The Madden Brothers? \\
      \midrule
      1. Which band is known for the singles "Chelsea Dagger" and "Whistle for the Choir"? \\
      - The Fratellis \\
      \\
      2. Which band had a top ten hit in the UK with "Chelsea Dagger"? \\
      - The Fratellis \\
      \\
      3. What was the highest chart position for "Whistle for the Choir" in the UK Singles Chart? \\
      - Number nine \\
      \\
      Answer: The Fratellis \\
      \bottomrule
    \end{tabular}
    \label{tab:fewshot_voting_example}
\end{table*}

\clearpage

\begin{table*}[ht]
    \caption{Prompt used for the CoT \cite{weiChainofThoughtPromptingElicits2023} experiments.}
    \centering
    \small
    \begin{tabular}{>{\raggedright\arraybackslash\tt}p{0.98\textwidth}<{}}
      \toprule
      You are a helpful question answerer who can provide an answer given a question and relevant context.\\
      Let's think through this step by step. Solve each step and explain how you arrived at your answer.\\
      Final answer should be a short span, just a few words.\\
      \bottomrule
    \end{tabular}
    \label{tab:cot_prompt}
\end{table*}

\begin{table*}
    \caption{The two-shot examples used for the CoT \cite{weiChainofThoughtPromptingElicits2023} configurations.}
    \centering
    \small
    \begin{tabular}{>{\raggedright\arraybackslash\tt}p{0.98\textwidth}<{}}
      \toprule
      Question: What is the shared profession of Jack Kerouac and Dan Masterson? \\
      \midrule
      Jack Kerouac and Dan Masterson are both writers. Jack Kerouac is best known as a novelist and a poet, a central figure of the Beat Generation. Dan Masterson is known as a poet. Therefore, their shared profession is writing, specifically in the realm of literature. \\
      \\
      Answer: they are poets. \\
      \midrule
      Question: Which band top the UK charts with their singles "Chelsea Dagger" and "Whistle for the Choir": The Fratellis or The Madden Brothers? \\
      \midrule
      The band that topped the UK charts with their singles "Chelsea Dagger" and "Whistle for the Choir" is The Fratellis. \\
      \\
      To determine this, we can look at the history and discography of each band: \\
      The Fratellis: \\
      "Chelsea Dagger" is a well-known single by The Fratellis, released in 2006 from their debut album "Costello Music." \\
      "Whistle for the Choir" is another single from the same album, also released in 2006.\\
      The Madden Brothers: \\
      This is a different musical duo consisting of Joel and Benji Madden from the band Good Charlotte. \\
      They do not have singles named "Chelsea Dagger" or "Whistle for the Choir."\\
      \\
      Given this information, it is clear that The Fratellis are the band associated with these singles.\\
      \\
      Answer: The Fratellis. \\
      \bottomrule
    \end{tabular}
    \label{tab:fewshot_cot_example}
\end{table*}

\begin{table*}[ht]
    \caption{Prompt used for the RaR \cite{dengRephraseRespondLet2024} experiments.}
    \centering
    \small
    \begin{tabular}{>{\raggedright\arraybackslash\tt}p{0.98\textwidth}<{}}
      \toprule
      You are a helpful question answerer who can provide an answer given a question and relevant context. \\
      Reframe the question with additional context and detail, then provide an answer.\\
      Final answer should be a short span, just a few words.\\
      \bottomrule
    \end{tabular}
    \label{tab:rar_prompt}
\end{table*}

\begin{table*}
    \caption{The two-shot examples used for the RaR \cite{dengRephraseRespondLet2024} configurations.}
    \centering
    \small
    \begin{tabular}{>{\raggedright\arraybackslash\tt}p{0.98\textwidth}<{}}
      \toprule
      Question: What is the shared profession of Jack Kerouac and Dan Masterson? \\
      \midrule
      What is the common profession that both Jack Kerouac, known for his influential work in the Beat Generation, and Dan Masterson, noted for his contributions to literature, share? \\
      \\
      Answer: Writers \\
      \midrule
      Question: Which band top the UK charts with their singles "Chelsea Dagger" and "Whistle for the Choir": The Fratellis or The Madden Brothers? \\
      \midrule
      Which band achieved the number one position on the UK charts with their hit singles "Chelsea Dagger" and "Whistle for the Choir": was it The Fratellis or The Madden Brothers? \\
      \\
      Answer: The Fratellis \\
      \bottomrule
    \end{tabular}
    \label{tab:fewshot_rar_example}
\end{table*}

\end{document}